\documentclass[10pt,twocolumn,letterpaper]{article}

\usepackage{iccv}
\usepackage{times}
\usepackage{epsfig}
\usepackage{graphicx}
\usepackage{amsmath}
\usepackage{amssymb}
\usepackage{algorithm} 
\usepackage{algorithmic}
\usepackage{helvet}
\usepackage[english]{babel}
\usepackage{courier}
\usepackage{bm}
\usepackage{indentfirst}
\usepackage{multirow}
\usepackage{blindtext}
\usepackage{color}
\usepackage{booktabs}
\usepackage{setspace}

\usepackage[pagebackref=true,breaklinks=true,letterpaper=true,colorlinks,bookmarks=false]{hyperref}
\makeatletter

\def\iccvPaperID{5853} 

\iccvfinalcopy
\begin{document}

\title{Foreground-aware Pyramid Reconstruction for \\Alignment-free Occluded Person Re-identification}

\author{Lingxiao He, Yinggang Wang, Wu Liu, Xingyu Liao, He Zhao, Zhenan Sun, Jiashi Feng\\
{\tt\small \{lingxiao.he\}@nlpr.ia.ac.cn}
}

\maketitle

\begin{abstract}
   Re-identifying a person across multiple disjoint camera views is important for intelligent video surveillance, smart retailing and many other applications. However, existing person re-identification (ReID) methods are challenged by the ubiquitous occlusion over persons and suffer from performance degradation. This paper proposes a novel occlusion-robust and alignment-free model for occluded person ReID and extends its application to realistic and crowded scenarios. The proposed model first leverages the full convolution network (FCN) and pyramid pooling to extract spatial pyramid features. Then an alignment-free matching approach, namely Foreground-aware Pyramid Reconstruction (FPR), is developed to accurately compute matching scores between occluded persons, despite their different scales and sizes. FPR uses the error from robust reconstruction over spatial pyramid features to measure similarities between two persons. More importantly, we design an occlusion-sensitive foreground probability generator that focuses more on clean human body parts to refine the similarity computation with less contamination from occlusion. The FPR is easily embedded into any end-to-end person ReID models. The effectiveness of the proposed method is clearly demonstrated by the experimental results (Rank-1 accuracy)  on three occluded person datasets: Partial REID (78.30\%), Partial iLIDS (68.08\%) and Occluded REID (81.00\%); and three benchmark person datasets: Market1501 (95.42\%), DukeMTMC (88.64\%) and CUHK03 (76.08\%).

\end{abstract}
\section{Introduction}
\begin{figure}[t]
    \centering
       \vspace{0em}
    \includegraphics[width=8.2cm]{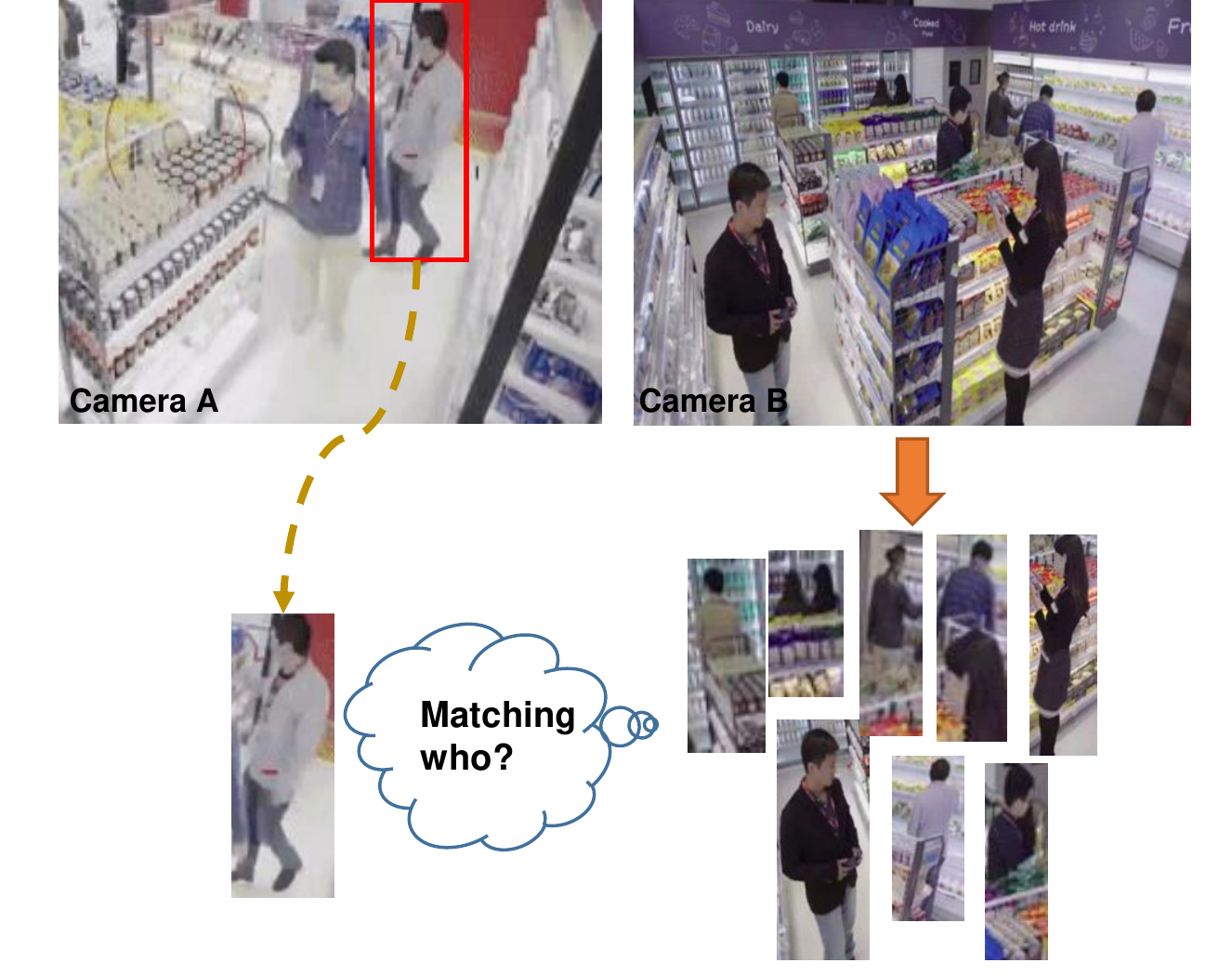}
     \caption{Illustration of the occluded person ReID problem. Here, the ReID system aims to recognize the person within the red bounding box  captured by camera A from several person images of different sizes captured by camera B. Most captured persons by the surveillance operator are occluded.}
    \label{fig:fig1}
\end{figure}

Person ReID is an important task with wide  real-world applications such as intelligent video surveillance, smart retailing, \emph{etc}., aiming at  matching person images captured from non-overlapping cameras.  One major issue that challenges this task is the ubiquitous occlusion over the captured persons. For example, as shown in Fig. \ref{fig:fig1}, people in an unmanned supermarket are occluded by goods,  shelves or other persons, making it difficult to track their movements.
Existing approaches \cite{kalayeh2018human, liu2018pose, qi2018maskreid, song2018mask} mostly leverage external cues, \emph{e.g.} person mask, semantic parsing or pose estimation, to align the detected persons. However, these approaches  may fail to generate accurate external cues in heavily occluded cases such as half body of a subject being occluded. Furthermore, it inevitably incurs more processing time to infer these external cues.  Some other approaches \cite{sun2017beyond, zheng2015partial}, by using part-based models, have achieved better  performance via part-to-part matching, but they require strict person alignment in advance.


In this paper, we propose a novel alignment-free approach that can re-identify persons accurately without requiring person alignment in advance even in the presence of heavy occlusion with the help of a foreground-aware pyramid reconstruction (FPR) based similarity measure. In particular, we firstly utilize the fully convolution network (FCN) to generate discriminative spatial feature maps that contain spatial coordinate information, and then post-process them via pyramid pooling, to extract spatial pyramid features. We then develop a novel matching score computation method that can be easily incorporated into any end-to-end person ReID model. More concretely, the proposed computation method encourages each spatial feature in the probe feature map to be linearly reconstructed from the basis spatial features within the gallery feature maps, and the average reconstruction error is used as the final matching score. In this way, the model is independent of the size of images and naturally skips the time-consuming alignment step. We also design a foreground probability generator to learn foreground probability maps (FPM) that can guide the spatial reconstruction by assigning the body parts with larger weights and the occlusion parts with smaller weights to overcome the occlusion problem. The proposed approach encourages the reconstruction error of the spatial feature maps extracted from the same person to be smaller than that of different identities. We conduct extensive experiments to validate the effectiveness of our proposed approach, and the results have clearly proved it can achieve accurate person ReID performance even in the presence of heavy occlusion.

To sum up, this work makes the following contributions:
\begin{itemize}
	
   \item We introduce a novel end-to-end spatial pyramid features learning architecture that can process input persons of different sizes and scales,  and generate discriminative features.
	
   \item We propose an occlusion-sensitive alignment-free approach, \emph{i.e.} foreground-aware pyramid reconstruction (FPR), that utilizes the foreground probability generator to guide the pyramid reconstruction for occluded person ReID. Unlike previous methods, it does not require any external cues in the application phase.

  \item  Experimental results demonstrate that the proposed approach achieves impressive results on multiple occlusion datasets including Partial REID \cite{zheng2015partial}, Partial iLIDS \cite{zheng2011person}, and Occluded REID \cite{zhuo2018}. It exceeds some occluded ReID approaches by more than 30\% in terms of Rank 1 accuracy. Additionally, FPR achieves competitive results on multiple benchmark person datasets including Market1501~\cite{zheng2015scalable}, DukeMTMC \cite{zheng2017unlabeled} and CUHK03~\cite{zheng2017pedestrian}.
\end{itemize}

\section{Related Work}
Occluded person ReID has attracted increasing attention due to its practical importance. Generally, previous methods for addressing this problem leverage external cues such as mask and pose, or adopt part-to-part matching.

\noindent\textbf{Approaches with External Cues.} \emph{Mask-guided models} \cite{kalayeh2018human, qi2018maskreid, song2018mask} use person masks that contain body shape information to help remove the background clutters at pixel-level for  person re-identification.  
For example, Kalayeh \emph{et al.} \cite{kalayeh2018human} proposed a  model that integrates human semantic parsing in person re-identification. It is similar to \cite{kalayeh2018human}, Qi \emph{et al.} \cite{qi2018maskreid} combined source images with person masks as the inputs to remove the appearance variations (illumination, pose, occlusion, etc.).
\emph{Pose-guided models} \cite{liu2018pose, su2017pose, suh2018part} utilize the skeleton as an external cue to effectively relieve the part misalignment problem by locating each part  using person landmarks. For instance, Su \emph{et al.} \cite{su2017pose} proposed a Pose-driven Deep Convolutional (PDC) model to learn improved feature extractors and matching models in an end-to-end manner. The PDC can explicitly leverage the human part cues to alleviate the identification difficulties caused by pose variations. Suh \emph{et al.} \cite{suh2018part} proposed a two-stream network, which consists of an appearance map extraction stream and a body part map extraction stream. Following the two streams, a part-aligned feature map was obtained by a bilinear mapping of the corresponding local appearance and body part descriptors. Although these approaches can indeed address occlusion problem, they heavily depend on accurate pedestrian segmentation, and also cost much time to infer the external cues.

\begin{figure*}[t]
    \centering
   \includegraphics[width=16.5cm]{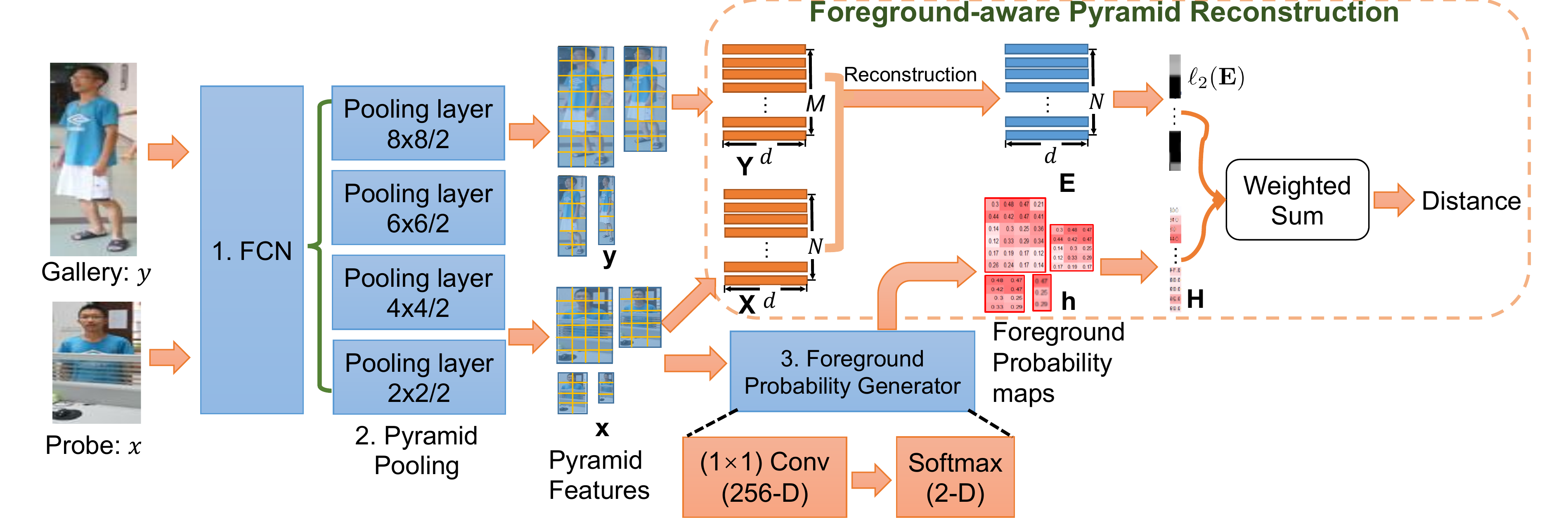}
     \caption{Architecture of the proposed foreground-aware pyramid reconstruction approach. It consists of three components: 1. a Fully Convolutional Network (FCN), 2. a Pyramid Pooling and 3. a Foreground Probability Generator. This structure can produce spatial pyramid features of inputs of different sizes and foreground probability maps $\mathbf{h}$. The second part is foreground-aware pyramid reconstruction for measuring the similarity between two person images. Given a probe $x$, the foreground probability vector $\mathbf{H}$ and spatial features $\mathbf{X}$ are obtained through foreground probability generator and FCN with Pyramid pooling respectively. Given gallery $y$, spatial features $\mathbf{Y}$ can be also obtained. Then we use linear reconstruction process to get the reconstruction error $\ell_2(\mathbf{E})$. Finally we perform weighted sum operation over $\ell_2(\mathbf{E})$ and $\mathbf{H}$ to obtain the similarity score between the probe $x$ and the gallery $y$.}
    \label{fig:fig2}
\end{figure*}
\begin{table}[t]
  \centering
  \small
  \caption{The comparison of occluded person ReID approaches along with the proposed FPR.}
  \label{tab110}
    \begin{tabular}{|l|c|c|}
    \hline
    \multirow{2}{*}{Approach}&Alignment&External cues\cr
     & requirement & requirement  \cr \hline
     Mask-guided& Require & Require\cr
     Pose-guided& Require & Require\cr
     Part-based& Require & Require \cr \hline
     FPR (ours)& Alignment-free & Do not require \cr\hline
    \end{tabular}
    \vspace{-0.5em}
\end{table}

\noindent\textbf{Part-based models}~\cite{sun2017beyond, wang2018learning, zhao2017spindle} employ a part-to-part matching strategy  to handle occlusion and mostly target at the cases where the person of interest is partially out of the camera's view. Zheng $\emph{et al.}$ \cite{zheng2015partial} proposed a local patch-level matching model called Ambiguity-sensitive Matching Classifier (AMC) based on dictionary learning with explicit patch ambiguity modeling, and introduced a global part-based matching model called Sliding Window Matching (SWM), which can provide complementary spatial layout information. However, the computation cost of AMC+SWM is rather expensive as features are calculated repeatedly without further acceleration. Sun \emph{et al.}~\cite{sun2017beyond} proposed a Part-based Convolutional Baseline (PCB) network that outputs a convolutional feature consisting of several part-level features. PCB focuses on the content consistency within each part to address the occlusion problem.  However, all these methods cannot skip the alignment step as well. He $\emph{et al.}$ \cite{he2018deep} proposed to reconstruct the feature map of holistic pedestrian from the visible parts by lasso regression for addressing partial person ReID.

Table \ref{tab110} compares the state-of-the-art algorithms to our approach about alignment and external cues requirement. It is noted that external cues based approaches are mainstream for occluded person ReID. However, accurate and stable external cues used for person alignment are hard to acquire in the application phase when half body is occluded. Different from previous approaches, our proposed method is alignment-free and more effective when it comes the ReID problem of occluded persons. It does not rely on any external cues while still achieves higher accuracy.

\section{Proposed Approach}
In this section, we elaborate on the proposed alignment-free occluded person re-identification approach. We first introduce the network architecture. After that, we introduce the foreground-aware pyramid reconstruction for computing matching scores between two persons with occlusion. Then we explain the training strategy of our model.

\subsection{Architecture of the Proposed Model}
The architecture of the proposed ReID model is shown in Fig. \ref{fig:fig2}. Structurally, it consists of a Full Convolutional Network (FCN), a Pyramid Pooling layer and a Foreground Probability Generator. We now explain them one by one.
\vspace{-1.2em}
\paragraph{FCN.} Conventional CNNs involving fully connected layers require a fixed-size input images as inputs. In fact, the requirement comes from fully-connected layers that demand fix-length vectors as inputs. Convolutional layers operate in a sliding-window manner and generate correspondingly-size spatial outputs. To handle an different sizes of person images, we discard all fully-connected layers to implement Fully Convolutional Network (FCN) that only convolution and pooling layers remain. Therefore, full convolutional network still retains spatial coordinate information, which is capable of extracting spatial features from different sizes of person images. The proposed FCN is based on ResNet-50 \cite{he2016deep}, it only contains 1 convolutional layer and 4 Resblocks layers, and the last Resblock outputs the spatial feature map.

\vspace{-1.2em}
\paragraph{Pyramid Pooling.} The detected persons for re-identification may have different scales, which makes it  difficult to align their spatial features  and brings errors to their similarity measure. To obtain robust spatial features regardless of scale variation, the features from FCN are further processed by a pyramid pooling layer to generate spatial pyramid features. The pyramid pooling layer consists of multiple max-pooling layers of different kernel sizes so that it has more comprehensive receptive fields over the input images. As shown in Fig. \ref{fig:fig2}, the output spatial features from the pooling layers of small kernel size capture the appearance information of a small local region. The output spatial features from the pooling layers of large kernel size capture the appearance information from relatively large regions in the image.  Finally, we concatenate the spatial pyramid features to obtain the final spatial feature that contains multi-scale information of the input thus the scale variation problem has been well addressed.
\vspace{-1.2em}
\paragraph{Foreground Probability Generator.} The target person to re-identify is provided with person detection bounding boxes. The detected person bounding boxes are coarse, often containing background and occlusion. Therefore, the output spatial features are contaminated by the occlusion and background. To guarantee the following spatial feature matching with less contamination from occlusion, we design a foreground probability generator to obtain the foreground probability maps (FPM). Such FPM would differentiate foreground from background and guide the following pyramid reconstruction for robust matching score computation. We will explain this module detailedly in the next subsection. As shown in Fig. \ref{fig:fig2}, the foreground probability map generator consists of a $1\times 1$ convolution layer and a softmax layer.


\subsection{Foreground-aware Pyramid Reconstruction}
Our proposed model performs foreground-aware pyramid reconstruction (FPR) to compute matching scores for input persons without requiring to align them in advance. Fig. \ref{fig:fig2} illustrates the workflow of FPR.

Suppose there is a pair of person images $x$ (probe: an occluded person image) and $y$ (gallery: an unoccluded person image), which may have different sizes. Denote the spatial pyramid maps of $x$ from FCN as $\mathbf{x} = [\mathbf{x}_k]_{k=1}^{K}$, where $\mathbf{x}$ consists of multi-scale feature maps generated from $K$ max-pooling layers in the pyramid pooling layer. Where $\mathbf{x}_k$ is a vectorized $w_x^{k}\times h_x^{k} \times d$ tensor, and $w_x^{k}$, $h_x^{k}$ and $d$ is the width, the height, the channel of the tensor. As shown in Fig. \ref{fig:fig2}, a total of $N$ spatial features from $N$ locations are aggregated into a matrix  $\mathbf{X}=[\mathbf{x}_n]_{n=1}^{N}\in \mathbb{R}^{d\times N}$, where $N= \sum_{k=1}^{K} w_x^{k}\times h_x^{k}$. Likewise, we construct the gallery feature matrix $\mathbf{Y}=[\mathbf{y}_m]_{m=1}^{M}\in \mathbb{R}^{d\times M}$, and $M =\sum_{k=1}^{K} w_y^{k}\times h_y^{k}$. Then, $\mathbf{x}_n$ that  denotes a local feature of a person part should be represented by a linear combination of $\mathbf{Y}$. In other words, some spatial features in $\mathbf{Y}$ should be able to linearly reconstruct $\mathbf{x}_n$ and the similarity between them can be computed as the reconstruction residual. Therefore, we first try to obtain the linear representation coefficients $\mathbf{w}_n$ of $\mathbf{x}_n$ with respect to $\mathbf{Y}$, where $\mathbf{w}_n \in \mathbb{R}^{N}$. With an $\ell_2$-norm regularization over $\mathbf{w}_n$, the linear representation formulation is
\begin{equation}
\begin{array}{l}
 \displaystyle \min_{\mathbf{w}_n}||\mathbf{x}_n-\mathbf{Y}\mathbf{w}_n||_2^{2}+\beta||\mathbf{w}_n||_2.
\end{array}
\label{eq1}
\end{equation}

For $N$ spatial features in $\mathbf{X}$, the Eq. (\ref{eq1}) can be rewritten as
\begin{equation}
\begin{array}{l}
 \displaystyle  \min_{\mathbf{W}}||\mathbf{X}-\mathbf{Y}\mathbf{W}||_2^{2}+\beta||\mathbf{W}||_F,
\end{array}
\label{eq4}
\end{equation}
where $\mathbf{W}=\{\mathbf{w}_1,\ldots,\mathbf{w}_N\}\in \mathbb{R}^{M\times N}$, and $\beta$ controls the smoothness of the coding vector $\mathbf{W}$.

We use the least square algorithm to solve $\mathbf{W}$, \emph{i.e.} $\mathbf{W}=(\mathbf{Y}^{T}\mathbf{Y}+\beta\cdot \mathbf{I})^{-1}\mathbf{Y}^{T}\mathbf{X}$. Then the reconstruction probe spatial features can be represented as
\begin{equation}
\begin{array}{l}
 \displaystyle \mathbf{\hat{X}} = \mathbf{Y}(\mathbf{Y}^{T}\mathbf{Y}+\beta\cdot \mathbf{I})^{-1}\mathbf{Y}^{T}\mathbf{X}.
\end{array}
\label{eq5}
\end{equation}
Let the residual spatial features $\mathbf{E} =\{\mathbf{E}_n\}_{n=1}^{N} = \mathbf{X}-\mathbf{\hat{X}}$. Then average reconstruction error is computed  as
\begin{equation}
\begin{array}{l}
 \displaystyle \text{distance} = \sum_{i=1}^{N}\ell_2(\mathbf{E})/N,
\end{array}
\label{eq6}
\end{equation}
where $\ell_2(\mathbf{E})=\{e_n\}_{n=1}^{N}\in \mathbb{R}^{1\times N}$, and $e_n$ is the spatial reconstruction error of the $n$-th spatial feature. The average reconstruction can be regarded as the distance between two person images.

With the above score computation, the alignment step in previous methods can be favourably avoided. However, it suffers from an obvious limitation: since the background and occlusion spatial features are all pooled into $\mathbf{X}$,  the reconstruction error of background or occlusion spatial features would be very large. As a consequence, the average reconstruction error increases, resulting in unreliable similarity scores and leads to mismatching. To address this problem, we propose to reduce the influence of background by assigning it small weights, while enhance the effect of foreground by assigning these regions large weights adaptively. Therefore, we consider using spatial foreground probability maps to guide spatial pyramid reconstruction for further obtaining the FPR model.

\begin{algorithm}[t]
\caption{Foreground-aware Pyramid Reconstruction (FPR)}
\label{alg:Framwork}
\begin{algorithmic}[1] 
\REQUIRE
A probe person image $x$; a gallery person image $y$.

\ENSURE Reconstruction error FPR. \\ 
\STATE Extract probe multi-scale spatial features $\mathbf{X}$ and multi-scale heatmaps $\mathbf{H}$ and gallery multi-scale spatial features $\mathbf{Y}$.
\STATE Solve Eq. (\ref{eq4}) to obtain reconstruction coefficient $\mathbf{W}$.
\STATE According to Eq. (\ref{eq5}) to calculate the reconstruction probe map $\hat{\mathbf{X}}$ to further to obtain residual map $\mathbf{E}$.
\STATE Solve Eq. (\ref{eq7}) to obtain the final FPR distance.
\end{algorithmic}
\end{algorithm}
Specifically, given the probe person image, the foreground probability generator as introduced above outputs spatial probability maps $\mathbf{h}$. Then the foreground probability vector $\mathbf{H} = [h_n]_{n=1}^{N}\in\mathbb{R}^{N}$ can be obtained, which reveals the different contributions of the spatial features from the probe image to spatial reconstruction. For the foreground spatial features, the output values in the FPM are relatively large, while for the background spatial features, the output values in the FPM are relatively small. Therefore, the ReID model can leverage the spatial vector $\mathbf{H}$ to guide the spatial reconstruction. We perform weighted sum operation over the reconstruction error $\ell_2(\mathbf{E})$ and the foreground probability vector $\mathbf{H}$. Then the FPR distance of two person images can be defined as 
\begin{equation}
\begin{array}{l}
 \displaystyle \text{distance} = \ell_2(\mathbf{E})*\mathbf{H}.
\end{array}
\label{eq7}
\end{equation}
The overall FPR is outlined in
Algorithm \ref{alg:Framwork}.

\begin{figure}[t]
    \centering
   \includegraphics[width=8cm]{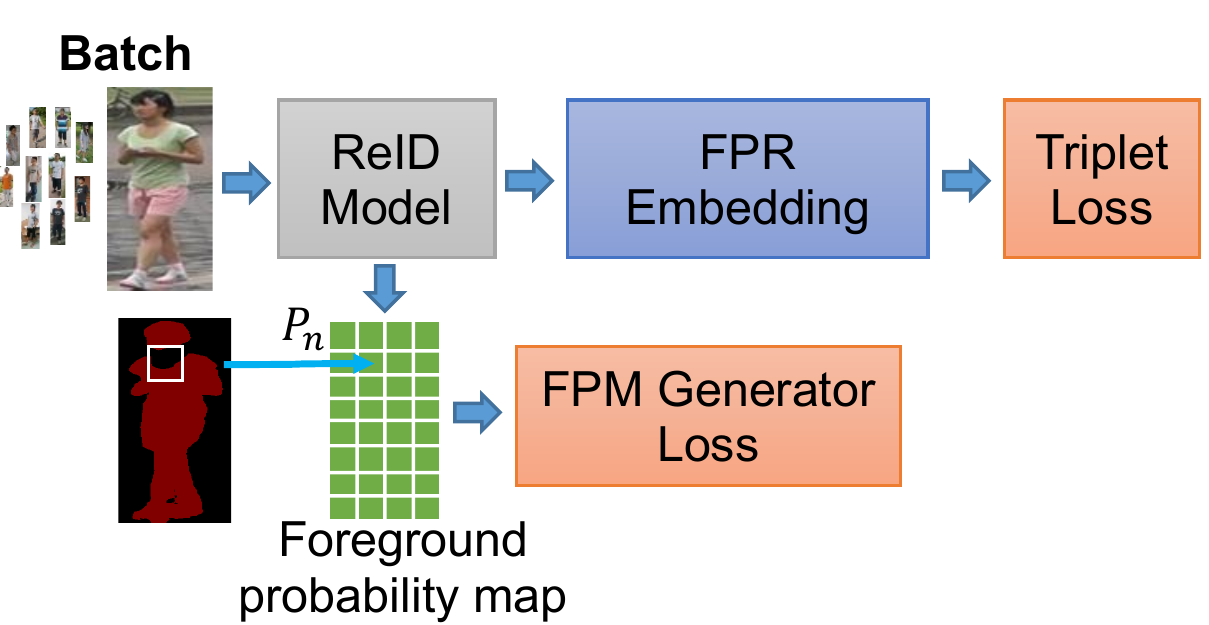}
     \caption{Model training. Our network consists of a batch of an input layer and a ReID model, in which FPR is embedded after the ReID network and followed by the triplet loss during training. Then the foreground probability generator loss learns the foreground probability map (FPM).}
    \label{fig:fig5}
\end{figure}

\subsection{Model Training}

We then explain  the training strategy of the foreground probability generator as well as the whole model.
Two loss functions, the triplet loss $\mathcal{L}_{\textup{tri}}$ and the foreground probability generator loss $\mathcal{L}_{\textup{fpg}}$ as shown in Fig.~\ref{fig:fig5}, are used to optimize the whole ReID model.

\vspace{-1.2em}
\paragraph{Triplet Loss} The  $\mathcal{L}_\textup{tri}$ is the hard example triplet loss function, which ensures that an image of a specific person is closer to all other images of the same person than any other images of a different person.

The goal of triplet embedding learning is to learn a function $f_{\theta}(x)$. Here, we want to make an image $x_i^{a}$ (anchor) of a specific person  closer to all other images $x_i^{p}$ (positive) of the same person than to any image $x_i^{n}$ (negative) of any other person in the image embedding space. Thus, we want $D(x_i^{a}, x_i^{p})+m<D(x_i^{a}, x_i^{n})$, where $D(:,:)$ is FPR measure between a pair of person images. Then the \emph{Triplet Loss} with $N$ samples is defined as  $ \sum_{i=1}^{N}[m+D(g_i^{a},g_i^{p})-D(g_i^{a},g_i^{n})]$, where $m$ is a margin that is enforced between a pair of positive and negative. To effectively select triple samples, the batch hard triplet loss modified by the triplet loss is adopted. The core idea  is to form batches by randomly sampling $P$ subjects, and then randomly sampling $K$ images of each subject, thus resulting in a batch of $PK$ images. Now, for each anchor sample in the batch, we can select the hardest positive and hardest negative samples within the batch when forming the triplets for computing the loss, which is called  the \emph{Batch Hard Triplet Loss}:
\begin{equation}
\begin{aligned}
 \displaystyle \mathcal{L}_{tri}(\theta)= \overbrace{\sum_{i=1}^{P}\sum_{a=1}^{K}}^{\text{all anchors}}[m&+\overbrace{\max_{p=1,\ldots,K}D({g}_i^{a},{g}_i^{p})}^{\text{hardest positive}}\\ & -\underbrace{\min_{n=1,\ldots,K}D({g}_i^{a},{g}_i^{n})}_{\text{hardest negative}}]
\end{aligned}
\end{equation}
\begin{figure}[t]
    \centering
    \vspace{0em}
    \includegraphics[width=6cm]{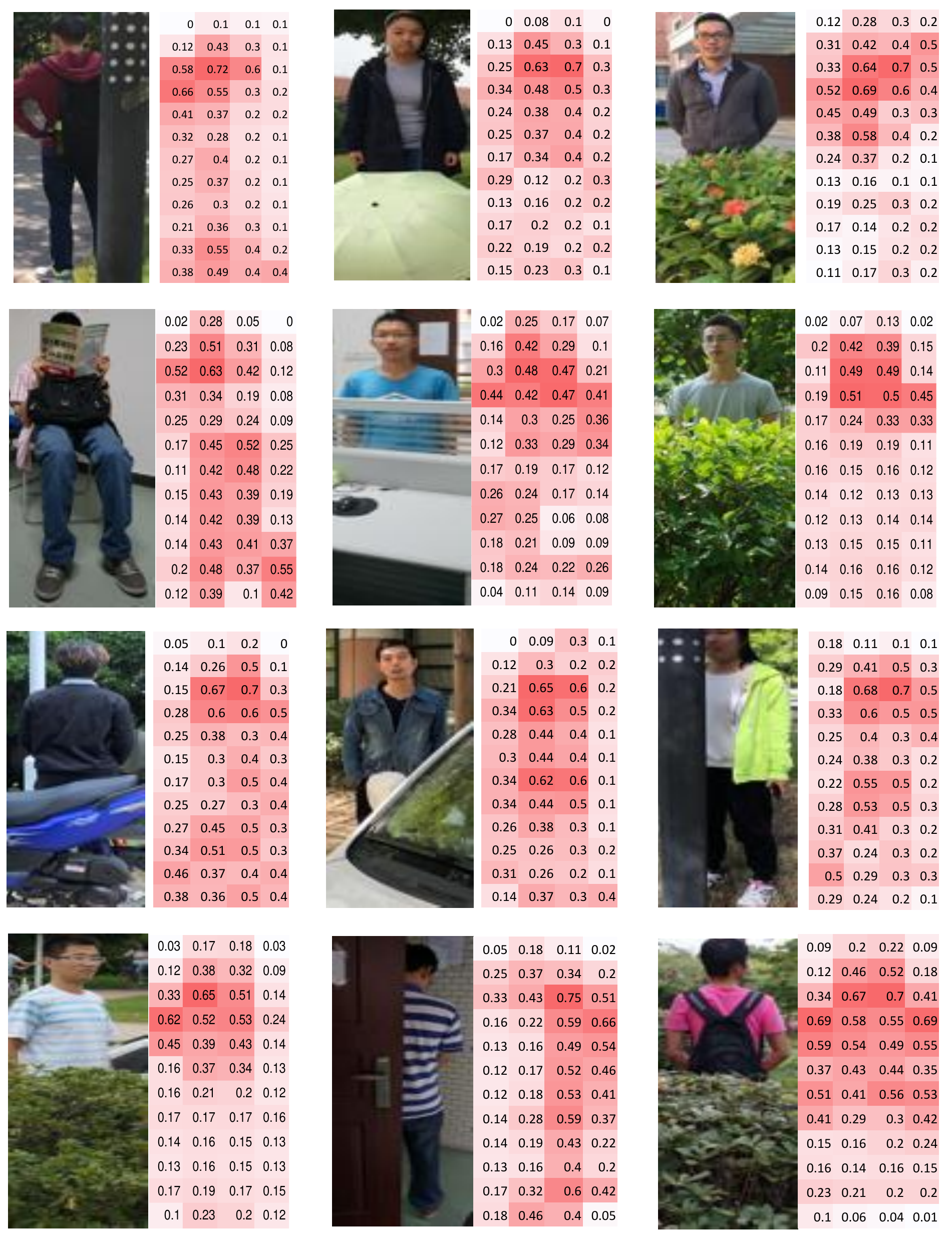}
    \caption{Foreground probability maps of occluded person images produced by foreground probability generator.}
    \label{fig:fig6}
\end{figure}
\vspace{-1.2em}
\paragraph{Foreground Probability Generator Loss} The $\mathcal{L}_\textup{fpg}$ is the spatial background-foreground classifier, which aims to classify the background/occlusion part and the person part.  We treat this problem as a binary classification problem. Given a person image, corresponding spatial features $\mathbf{X} = \{\mathbf{x}_n\}_{n=1}^{N}$ are extracted. The label of $\mathbf{x}_n$ is determined by the person mask obtained by the semantic segmentation model \cite{liu2018devil}. The spatial feature $\mathbf{x}_n$ corresponds to the mask region $P_n$. We calculate the average pixel value of $P_n$ to obtain its mask-label $m_n$:
\begin{equation}
\begin{array}{l}
\displaystyle m_n = \frac{\sum_{w=1}^W\sum_{h=1}^{H}P_i^{w,h}}{W\times H},
\end{array}
\end{equation}
where $W, H$ are the width and the height of the mask patch $P_n$. Then we set a label threshold $\tau$ ($0\leq \tau \leq 1$) to obtain the labels of spatial features.  The spatial background/foreground label can be defined as
\begin{equation}
y_n = \left\{
\begin{array}{lr}
0,\,\, m_n\leq\tau  \\
1,\,\, m_n>\tau ,\\
\end{array}
\label{eq12}
\right.
\end{equation}
where $\tau$ is the label threshold and $0\leq \tau \leq 1$. The foreground probability generator loss function is then given by
\begin{equation}
\begin{array}{l}
\displaystyle \mathcal{L}_{\textup{fpg}} = \sum_{n=1}^{N}[y_n \log(f_{\theta}(\mathbf{x}_n))+(1-y_n)log(1- f_{\theta}(\mathbf{x}_n))],
\end{array}
\end{equation}
where $y_n = 0$ and $y_n=1$ respectively indicate the background and foreground spatial feature labels.

Fig. \ref{fig:fig6} shows some FPM of occluded person images that are generated by the softmax layer. We can see that the spatial background-foreground classifier can accurately detect the person parts.

The final total loss function is defined as
\begin{equation}
\begin{array}{l}
\displaystyle \mathcal{L}_{total} = \mathcal{L}_{\textup{tri}} + \alpha\mathcal{L}_{\textup{bfc}},
\end{array}
\label{eq11}
\end{equation}
where $\alpha$ controls the importance of the spatial foreground probability generator loss function.

\section{Experiments}
In this section we first verify the effectiveness of our proposed approach for the task of  occluded person re-identification, and then experiment on non-occluded datasets to test its generalizability. Also, we perform parameter analysis  to investigate the  influence of weight $\alpha$ and threshold $\tau$ in training and testing phases.

\subsection{Experiment Settings}
\noindent\textbf{Implementation Details.}
Our implementation is based on the publicly available code of PyTorch. All models are trained and tested on Linux with GTX TITAN X GPUs. During training, all training samples are all re-scaled to $384\times 128$. No data augmentation is used. Besides, we empirically set $\alpha$ = 0.02 in Eq. (\ref{eq11}), $\tau$ = 0.35 in Eq. (\ref{eq12}) and $\beta = 0.01$ in Eq. (\ref{eq4}). For the batch hard triplet loss function, one batch consists of 16 subjects, and each subject has 4 different images. Therefore, each batch returns 64 groups of hard triples. The  proposed model is trained with 200 epochs.

\noindent\textbf{Evaluation Protocol.} For performance evaluation, we employ the standard metrics as in most person ReID literature, namely the cumulative matching cure (CMC) and the mean Average Precision (mAP). To evaluate our method, we re-implement the evaluation code provided by \cite{zheng2015scalable} in Python.

\begin{table}[t]
  \centering
  \small
  \caption{Databases used in the occluded person ReID experiments. Market1501 dataset is used for training the ReID model, and the three occluded person datasets used for testing.}
  \label{tab1}
    \begin{tabular}{|l|c|c|c|}
    \hline
    \multirow{2}{*}{Database} &
   {Training}&\multicolumn{2}{c|}{Testing (\#id/\#imgs)} \cr \cline{3-4}
     & (\#id/\#imgs)&Gallery & Probe\cr \hline
     \textbf{Partial REID}   &-&60/300&60/300\cr
     \textbf{Partial iLIDS}  &-&119/238&119/238\cr
     \textbf{Occluded REID}   &-&200/1,000&200/1,000\cr \hline
    \end{tabular}
\end{table}
\begin{figure}[t]
    \centering
       \vspace{0em}
    \includegraphics[width=8cm]{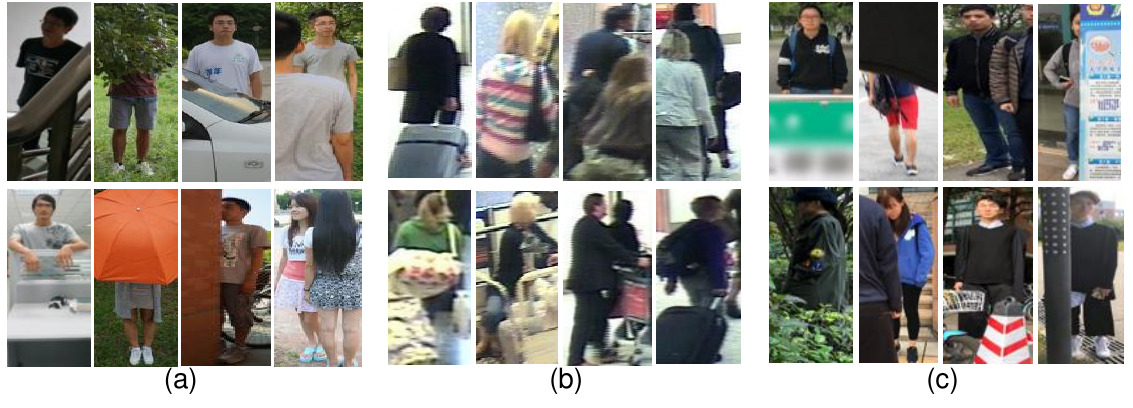}
     \caption{Examples of occluded persons in (a) Partial REID, (b) Partial iLIDS, and (c) Occluded REID  datasets. }
    \label{fig:fig7}
    \vspace{-0.5em}
\end{figure}
\subsection{Evaluation on Occluded Person Datasets}
\noindent\textbf{Datasets.}
\emph{Partial REID} \cite{zheng2015partial} is a specially designed partial person dataset that includes 600 images from 60 people, with 5 full-body images and 5 occluded images per person. These images were collected on a university campus by 6 cameras from different viewpoints, backgrounds and different types of occlusion. The examples of partial persons in the Partial REID dataset are shown in Fig. \ref{fig:fig7}(a). We follow the evaluation protocols in \cite{zheng2015scalable} where 300 full-body images of 60 identities are used as the  gallery set and 300 occluded-body images of the same 60 identities are used as the probe set. \emph{Partial iLIDS} \cite{he2018deep} contains a total of 476 images of 119 people captured by 4 non-overlapping cameras. Some images contain people occluded by other individuals or luggage. Fig. \ref{fig:fig7}(b) shows some examples of individual images from the iLIDS dataset. For the gallery set,
238 images of 119 individuals captured by 1st, 2nd cameras are used as the  gallery set and 238 images of 119 individuals captured 3rd, 4th cameras are used as a probe set. \emph{Occluded REID} \cite{zhuo2018} is an occluded person dataset captured by mobile cameras, consisting of 2,000 images of 200 occluded persons (see Fig. \ref{fig:fig7}(c)). Each identity has 5 full-body person images and 5 occluded person images with different types of occlusion. All images with different viewpoints and backgrounds are resized to $384\times 128$. The details of the training set and testing set are shown in Table \ref{tab1}.

\begin{table}[t]
\small
  \centering
    \caption{Performance comparison on Partial REID, Partial-iLIDS and Occluded REID datasets. R1: rank-1. mAP: mean Accuracy Precision.}
    \begin{tabular}{|l|cc|}
    \hline
    &
    \multicolumn{2}{c|}{\textbf{Occluded REID}}\cr\cline{2-3}
     &R1&mAP \cr
   \hline
    MaskReID \cite{qi2018maskreid}&26.80&25.00 \cr
    PCB \cite{sun2017beyond}&41.30&38.90 \cr
    AMC+SWM \cite{zheng2015partial}&31.12&27.33\cr
    DSR \cite{he2018deep}& 72.80& 62.83 \cr\hline
    Baseline & 42.12&37.24 \cr

    FPR (ours)& \bf 78.30& \bf 68.00 \cr\hline
    \end{tabular}
    \label{tab2}
\end{table}
\renewcommand{\arraystretch}{1.15}
\begin{table}[t]
\small
  \centering
    \begin{tabular}{|l|cc|cc|}
    \hline
    &
    \multicolumn{2}{c|}{\textbf{Partial REID}}&\multicolumn{2}{c|}{\textbf{Partial iLIDS}}\cr\cline{2-5}
     &R1&mAP&R1&mAP \cr
   \hline
    MaskReID \cite{qi2018maskreid}&28.70&32.20&33.00&30.40 \cr
    PCB \cite{sun2017beyond} &56.30&54.70&46.80&40.20 \cr
    AMC+SWM \cite{zheng2015partial}&34.27&31.33&38.67&31.33\cr
    DSR \cite{he2018deep}& 73.67& 68.07 &64.29&58.12\cr\hline
    Baseline & 53.33&50.20 &52.94&43.53\cr
    FPR (ours)& \bf 81.00& \bf 76.60 & \bf 68.08& \bf 61.78\cr\hline
    \end{tabular}
    \label{tab112}
\end{table}

\begin{table*}[t]
  \centering
  \small
  \caption{Performance comparison on Market1501, CHUK03 and DukeMTMC datasets. R1: rank-1. mAP: mean Accuracy Precision. }
  \label{tab3}
    \begin{tabular}{|ll|cc|cc|cc|}
    \hline
    \multicolumn{2}{|l}{\multirow{2}{*}{Method}} &
    \multicolumn{2}{|c|}{\textbf{Market1501}}&\multicolumn{2}{|c|}{\textbf{CUHK03}}&\multicolumn{2}{|c|}{\textbf{DukeMTMC}} \cr \cline{3-8}
     \multicolumn{2}{|c|}{~}& R1 &mAP &R1 &mAP &R1 &mAP   \cr \hline
     \multirow{3}{*}{Part-based}
    &PCB (ECCV18) \cite{sun2017beyond}&92.30&77.40&61.30&54.20&81.80&66.10\cr
    &PCB+RPP (ECCV18) \cite{sun2017beyond}&93.80&81.60&63.70&57.50&83.30&69.20\cr
    & {DSR (CVPR18) \cite{he2018deep}}&94.71&85.78&75.24&71.15&88.14&77.07 \cr\hline
     \multirow{3}{*}{Mask-guided}  &SPReID (CVPR18) \cite{kalayeh2018human}& 92.54& 81.34&-&\-&-&-\cr
    &MGCAM (CVPR18) \cite{song2018mask}&83.79 &74.33&50.14 &50.21&46.71&46.87 \cr
    &MaskReID (Arxiv18) \cite{qi2018maskreid} & 90.02 &75.30 & -&- &- &- \cr \hline
    \multirow{4}{*}{Pose-guided}  &PDC (ICCV17) \cite{su2017pose}&84.14&63.41&-&-&-&- \cr
    &PABR (Arxiv18) \cite{suh2018part}&90.20&76.00&-&-&-&-\cr
    &Pose-transfer (CVPR18) \cite{liu2018pose}&87.65&68.92&33.80&30.50&30.10&28.20 \cr
    &PSE (CVPR18) \cite{sarfraz2017pose}&87.70&69.00&-&-&27.30&30.20\cr \hline
    \multirow{3}{*}{Attention-based}  &DuATM (CVPR18) \cite{si2018dual}&91.42&76.62&-&-&-&- \cr
    &HA-CNN (CVPR18) \cite{li2018harmonious}&91.20&75.70&44.40&41.00&41.70&38.60\cr
    &AACN (CVPR18) \cite{xu2018attention}&85.90&66.87&-&-&-&- \cr \hline
    \multicolumn{2}{|l|}{Baseline}&94.06&84.62&73.57&69.35&87.30&76.18 \cr

   \multicolumn{2}{|l|}{FPR (ours)}&\bf 95.42& \bf 86.58&\bf 76.08 & \bf 72.31&\bf 88.64&  \bf 78.42\cr\hline
    \end{tabular}

\end{table*}
\noindent\textbf{Benchmark Algorithms.} Several existing partial person ReID methods are used for comparison, including Ambiguity-sensitive Matching (AMC) with Sliding Window Matching (SWM) \cite{zheng2015partial} (AMC + SWM), PCB \cite{sun2017beyond} and DSR \cite{he2018deep},  which are two part-based matching methods; and the mask-guided ReID model MaskReID \cite{qi2018maskreid}. For AMC + SWM, features are extracted from $32\times 32$ supporting areas which are densely sampled with an overlap of half of the height/width of the supporting area in both horizontal and vertical directions. Each region is represented following \cite{zheng2015partial}. Besides, the weights of AMC and SWM are 0.7 and 0.3, respectively. For PCB and MaskReID, we follow their original parameter settings. Our ReID model is trained with Market1501. We follow the standard training protocols in \cite{zheng2015scalable}, where 751 identities are used for training. Therefore, it is also a cross-domain setting.

\begin{figure}[t]
    \centering
    \includegraphics[width=8.2cm]{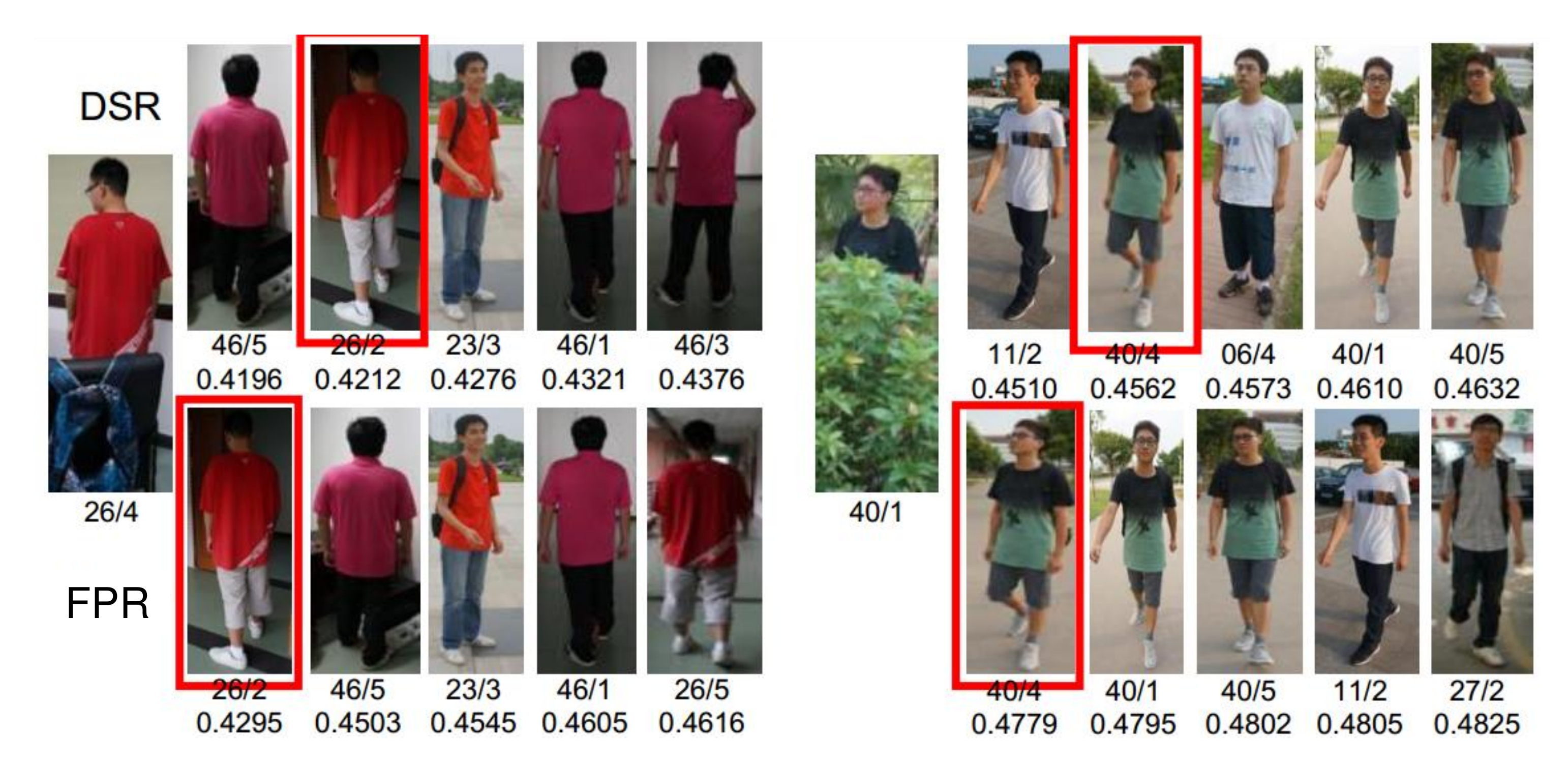}
     \caption{Occluded person retrieval of DSR and FPR. The red bounding indicates the correct retrieval result, we find that FPR can address the case where DSR cannot get the correct result with smaller reconstruction error.}
    \label{fig:fig8}
    \vspace{-0.5em}
\end{figure}
\noindent\textbf{Results.} Table \ref{tab2} shows the experimental results. We find the results on Partial REID, Partial iLIDS and occluded REID are similar. The proposed method FPR outperforms MaskReID, PCB, AMC-SWM and DSR with R1 76.33\%, 68.07\% and 76.30\% and mAP 76.60\%, 61.78\%, 68.00\% respectively on the three occluded person datasets. Note that the gap between FPR and DSR is significant. Our method FPR increases R1 Accuracy from 73.67\% to 81.00\%, from 64.29\% to 68.07\%, and from 72.80\% to 78.30\% on the three occluded person datasets respectively.  This is because background and occlusion largely affect reconstruction error, and then lead to larger average error.
Remarkably, FPR effectively reduces the influence of background and occlusion by assigning them small weights. For these comparison approaches, PCB is unable to relieve the influence of occlusion and background since it fuses both occlusion/background part feature and human part feature to the final feature. Although MaskReID is well suited for addressing person occlusion problem, it depends on external cues such as masks during the inference. The proposed FPR is an alignment-free approach since it does not depend on external cues to align the person images. The retrieval results are shown in Fig. \ref{fig:fig8}.  Experiments are conducted using the cross-domain setting and no images in the three partial datasets are used for training (Market1501 training set is used to obtain the ReID model). The FPR achieves good cross-domain performance in comparison with other approaches.

\subsection{Evaluation on Non-occluded Person Datasets}
We also experiment on non-occluded person datasets to test the generalizability of our proposed approach.

\begin{table}[t]
  \centering
  \small
  \caption{Databases used in the unoccluded person ReID experiments.}
  \label{tab22}
    \begin{tabular}{|l|c|c|c|}
    \hline
    \multirow{2}{*}{Database} &
   {Training}&\multicolumn{2}{c|}{Testing (\#id/\#imgs)} \cr \cline{3-4}
     & (\#id/\#imgs)&Gallery & Probe\cr \hline
   \textbf{ Market1501} &751/12,936&750/15,913&750/3,368 \cr
    \textbf{DukeMTMC}  &702/16,522&1,110/17,661&702/2,228 \cr
    \textbf{CUHK03}  &767/7,365&700/5,332&700/1,400 \cr \hline
    \end{tabular}
\end{table}

\noindent\textbf{Datasets.} Three person re-identification datasets \emph{Market1501} \cite{zheng2015scalable}, \emph{CUHK03} \cite{zheng2017pedestrian} and \emph{DukeMTMC-reID} \cite{zheng2017unlabeled} are used. Market1501 has 12,936 training and 19,732 testing images with 1,501 identities in total from 6 cameras. Deformable Part Model (DPM) is used as the person detector. We follow the standard training and evaluation protocols in \cite{zheng2015scalable} where 751 identities are used for training and the remaining 750 identities for testing. CUHK03 consists of 13,164 images of 1,467 subjects captured by 2 cameras on CUHK campus. Both manually labelled and DFM detected person bounding boxes are provided. We adopt the new training/testing protocol \cite{zheng2017pedestrian} proposed since it defines a more realistic and challenging ReID task. In particular, 767 identities are used for training and the remaining 700 identities are used for testing. DukeMTMC-reID is the subset of Duke Dataset \cite{ristani2016performance}, which consists of 16,522 training images from 702 identities, 2,228 query images and 17,661 gallery images from the other identities. It provides manually labelled person bounding boxes. Here, we follow the setup in \cite{zheng2017unlabeled}. The details of training and testing sets are shown in Table \ref{tab22}.

\noindent\textbf{Results.}
Comparisons are made between FPR and 10 state-of-the-art approaches of four categories, including part-based model: PCB \cite{sun2017beyond}, mask-guided models: SPReID \cite{kalayeh2018human}, MGCAM \cite{song2018mask}, MaskReID \cite{qi2018maskreid}, pose-guided models: PDC \cite{su2017pose}, PABR \cite{suh2018part}, Pose-transfer \cite{liu2018pose}, PSE \cite{sarfraz2017pose} and attention-based models: DuATM \cite{si2018dual}, HA-CNN \cite{li2018harmonious}, AACN \cite{xu2018attention}, on Market1501, CUHK03, DukeMTMC datasets. The results  are shown in Table \ref{tab3}. From the table, it can be seen that  the proposed FPR achieves competitive performance for all evaluations.

The gaps between FRP and DSR are significant. FPR increases R1 Accuracy from 94.71\% to 95.42\%, from 75.24\% to 76.08\%, from 88.14\% to 88.64\% on Market1501, CUHK03 and DukeMTMC, respectively. FPR increases mAP from 85.78\% to 86.58\%, from 71.15\% to 72.31\%, from 77.07\% to 78.42\% on Market1501, CUHK03 and DukeMTMC, respectively. These results demonstrate that the designed foreground probability generator in deep spatial reconstruction is very useful. Besides, FPR performs better than part-based model PCB, because part-level features cannot eliminate the impact of occlusion and background. Furthermore, the proposed FPR is superior to some approaches with external cues. The mask-guided and pose-guided approaches heavily rely on the external cues for person alignment, but they cannot always infer the accurate external cues in the case of severe occlusion, thus resulting in mismatching. FPR utilizes foreground probability maps to guide spatial reconstruction, which naturally avoid alignment and can address person images even in presence of heavy occlusion. Not only do the proposed FPR get good performance at R1 accuracy, but also it is superior to other methods at mAP.


\begin{figure}[t]
    \centering
    \includegraphics[width=8.8cm]{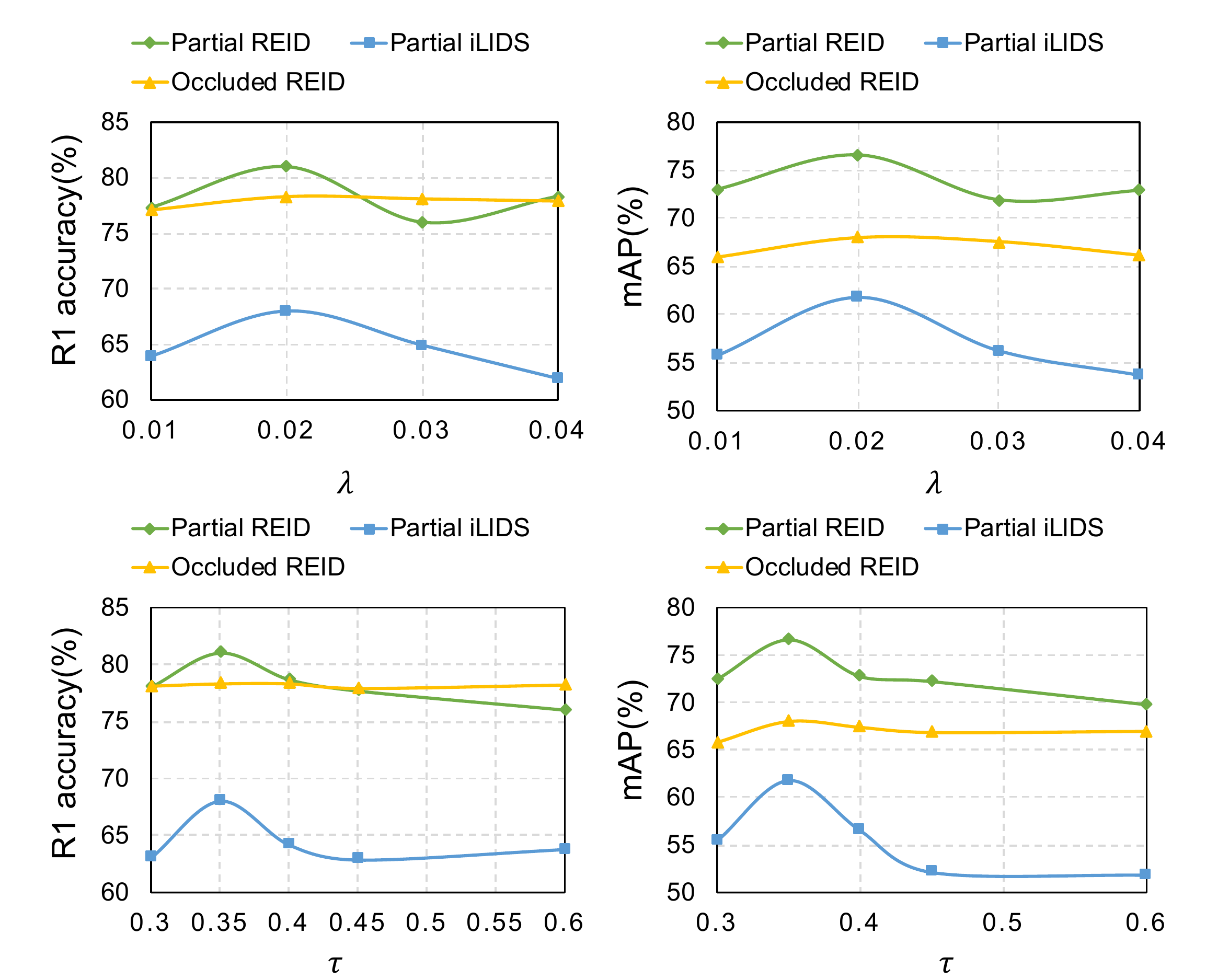}
     \caption{Evaluation of different parameters of FPR (Eq. (\ref{eq12})$\&$(\ref{eq11})) using Rank-1 and mAP accuracy on the three occluded datasets.}
    \label{fig:fig10}
\end{figure}

\subsection{Parameter Analysis}
We evaluate two key parameters in our modelling, the label threshold $\tau$ in Eq. (\ref{eq11}) and the weight $\alpha$ of spatial foreground probability generator loss in Eq. (\ref{eq12}). The two parameters would influence the performance of the proposed FPR. To explore the influence of $\alpha$ to FPR, we fix $\tau=0.35$ and set the value of $\alpha$ from 0.01 to 0.04 at the stride of 0.01. We show the results on the three occluded person datasets in Fig. \ref{fig:fig10}, we find that the proposed FPR achieves the best performance when we set $\alpha=0.02$. To further explore the influence of $\tau$ to FPR, we fix $\alpha=0.02$ and set the value of $\tau$ from 0 to 1 at the stride of 0.1. As shown in Fig. \ref{fig:fig10}, when $\tau$ is approximately 0.35, the proposed FPR achieves the best performance.



\section{Conclusions}
We have proposed a novel and alignment-free approach called Foreground-aware Pyramid Reconstruction (FPR) to occluded person ReID. The proposed method provides a feasible scheme where the probe spatial feature can be linearly reconstructed by gallery spatial featuresto achieve effective alignment-free matching. More importantly, spatial foreground probability used in the reconstruction process can fully solve the occlusion problem. Furthermore, we embedded FPR into batch hard triplet loss function to learn more discriminative features by minimizing the reconstruction error for an image pair from the same target and maximizing that of image pair from different targets. Experimental results on three occluded datasets validate the effectiveness of FPR. Additionally, the proposed method is also competitive on the benchmark person datasets. \\


{
\bibliographystyle{ieee}
\bibliography{egbib}
}

\end{document}